\documentclass{bmvc2k}

\title{Flow-based GAN for 3D Point Cloud Generation from a Single Image}

\addauthor{Yao Wei}{yao.wei@utwente.nl}{1}
\addauthor{George Vosselman}{george.vosselman@utwente.nl}{1}
\addauthor{Michael Ying Yang}{michael.yang@utwente.nl}{1}

\addinstitution{
 University of Twente\\
 The Netherlands
}

\runninghead{Wei, Vosselman, Yang}{Flow-based GAN for 3D Point Cloud Generation}

\usepackage{amsfonts, amssymb}
\usepackage{multirow}
\DeclareUnicodeCharacter{2212}{-}
\usepackage{url}

\usepackage{enumitem}
\setitemize{noitemsep,topsep=0pt,parsep=0pt,partopsep=0pt}

\usepackage{soul, color, xcolor}

\begin{document}

\maketitle

\begin{abstract}
Generating a 3D point cloud from a single 2D image is of great importance for 3D scene understanding applications. To reconstruct the whole 3D shape of the object shown in the image, the existing deep learning based approaches use either explicit or implicit generative modeling of point clouds, which, however, suffer from limited quality. In this work, we aim to alleviate this issue by introducing a hybrid explicit-implicit generative modeling scheme, which inherits the flow-based explicit generative models for sampling point clouds with arbitrary resolutions while improving the detailed 3D structures of point clouds by leveraging the implicit generative adversarial networks (GANs). We evaluate on the large-scale synthetic dataset ShapeNet, with the experimental results demonstrating the superior performance of the proposed method. In addition, the generalization ability of our method is demonstrated by performing on cross-category synthetic images as well as by testing on real images from PASCAL3D+ dataset. Code available at: \url{https://github.com/weiyao1996/FlowGAN}.
\end{abstract}

\section{Introduction}
\label{sec:intro}

As a popular representation for modeling the 3D shapes of objects, point clouds play an important role in 3D scene understanding applications including autonomous driving, augmented reality (AR) and virtual reality (VR). Typically, 3D point clouds can be acquired by laser scanning techniques (e.g., LiDAR), or reconstructed from multi-view images by SfM. However, it is costly to obtain satisfactory point clouds. For instance, LiDAR sensors are more expensive than regular cameras, and the limited sensor resolution may result in sparse and incomplete raw point clouds. In addition, SfM techniques are sometimes not practical, and feature matching becomes hard when multiple viewpoints are separated by a large margin and local appearance varies a lot. Therefore, the efficient production of point clouds could be a concern in 3D computer vision.

\begin{figure}[h]
    \centering
    \includegraphics[width=.79\linewidth]{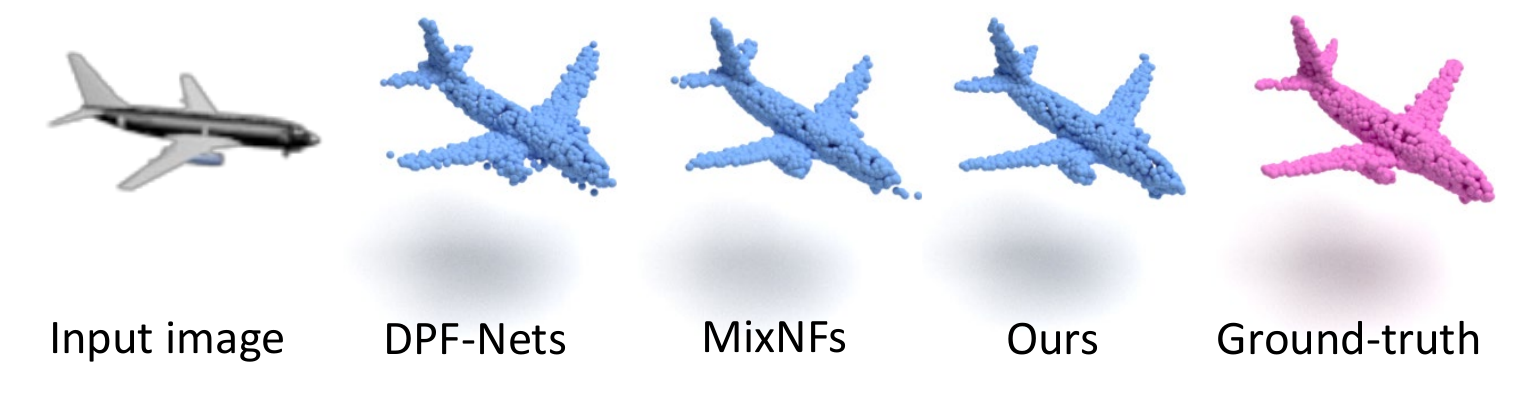}
    \caption{Unlike early methods which suffer from fixed resolutions, our method (Ours) inherits the flexibility of flow-based methods for generating an arbitrary number of points. Moreover, our single-flow-based method can generate point clouds with cleaner global shapes and finer details while reducing computation costs during inference, compared to recent flow-based methods DPF-Nets \cite{klokov2020discrete} and MixNFs \cite{postels2021go} which rely on multiple flow models.}
    \vspace{-5mm}
    \label{fig:Preview}
\end{figure}

Motivated by the recent success of deep generative models in the 2D domain \cite{karras2019style}, data generation has attracted increasing attention in the 3D domain. Deep generative models benefit from deep learning and generative modeling, which exploit deep neural networks to model the probability distribution (i.e., density function) of training data, thereby new samples that follow such density function can be generated. Generative models can generally be divided into two categories \cite{goodfellow2016nips}: explicit density models (e.g., Variational Autoencoders (VAEs) \cite{kingma2013auto} and Normalizing Flows (NFs) \cite{rezende2015variational}), and implicit density models (e.g., Generative Adversarial Networks (GANs) \cite{goodfellow2014generative}). The former explicitly defines and approximates the density function while the latter implicitly represents the density function by sampling from it. With the advances in these deep generative models, various approaches for 3D point cloud generation \cite{fan2017point, achlioptas2018learning, yang2019pointflow, sun2020pointgrow, luo2021diffusion} have been proposed and made great progress. 

However, most of the existing point cloud generation methods only utilize one (either explicit or implicit) generative modeling, leading to limitations especially in terms of flexibility and efficiency. By employing VAEs or GANs, early works \cite{fan2017point, achlioptas2018learning, gadelha2018multiresolution, groueix2018papier, mandikal20183d, wang2018pixel2mesh, wang2019deep} regard a point cloud as a set of 3D coordinates organized in a $N\times3$ matrix, where $N$ is fixed and pre-defined, referred to as the point cloud resolution. These methods usually adopt point-set distances as optimization objectives, however, only generate point clouds with a fixed resolution which limits the ability of producing complex 3D shapes \cite{yang2019pointflow}. Alternatively, some NFs based works \cite{yang2019pointflow, klokov2020discrete} treat point clouds as sampling from a distribution of 3D points, which allows generating point clouds with an arbitrary number of points, however, these methods may introduce more noise and fail to maintain detailed 3D structures \cite{postels2021go}. 

The goal of this work is to develop a new generative modeling scheme to reconstruct 3D point clouds given a single input 2D image, which can be referred to as point cloud single-view reconstruction. As mentioned above, previous works which only consider either explicit or implicit generative modeling of point clouds, suffer from fixed resolution and/or coarse detail issues. To address these issues, we propose a hybrid explicit-implicit generative modeling scheme, which enables flexible and efficient point cloud reconstruction from a single image. Technically, a generative adversarial framework containing a generator and a discriminator is introduced to perform hybrid explicit-implicit generative modeling. 
The \textbf{main contributions} are summarized as follows:
\begin{itemize}
    \item We propose a new generative modeling scheme for point cloud reconstruction, which inherits the flow-based generative methods for generating an arbitrary number of points, while improving the detailed 3D structures of the generated point clouds by leveraging an adversarial training strategy.
    \item A simple and effective discriminator is introduced to guide a flow-based generator to reconstruct high-quality 3D point clouds.  
    \item Experimental results demonstrate that the proposed method achieves state-of-the-art performance in 3D point cloud reconstruction from a single image.
\end{itemize}

\section{Related Work}
\vspace{-3mm}
\subsection{Deep Generative Models for 3D Point Cloud}

Given a variable $a$ with a set of observations $\{a_1, a_2, …\}$, a generative model learns to estimate the probability density function $p(a)$, which can be utilized to generate new samples from the underlying distribution. By aggregating the benefits of deep learning and generative modeling, deep generative models have shown great potential for data synthesis in both 2D and 3D space, and have particularly achieved success in image synthesis. Different from structured grid-like images, 3D point clouds are irregular unordered point sets, making it difficult to directly apply current image generation approaches to point cloud generation. The existing point cloud generation approaches explore different learning representations of point clouds, which can be roughly classified into two categories, i.e., point-based and probabilistic distribution-based. While remarkable progress has been made, these two types of methods have their inherent limitations.

\noindent
\textbf{Point-based:} Early methods typically represent a point cloud as a fixed dimensional matrix (e.g., $N\times3$ containing the 3D coordinates of $N$ points), and adopt VAEs \cite{gadelha2018multiresolution} or GANs \cite{achlioptas2018learning} to predict point clouds from certain inputs, where the models are optimized by minimizing Chamfer distance (CD) or Earth Mover’s distance (EMD) between predicted point clouds and ground-truth point clouds. The main drawbacks include: generating point clouds with a fixed number of points greatly limits the ability in producing complex 3D shapes; those point-set distances are far from ideal objectives. For example, CD favors point clouds that are concentrated in the mode of the marginal point distribution; and EMD is often computed by approximations, which may be computational expensive and lead to biased gradients.

\noindent
\textbf{Probabilistic distribution-based:} Alternatively, point clouds can be viewed as samples from a probabilistic distribution of 3D points, which inspires exploration on applying flow models which consist of a series of invertible transformation. The highlights lie in the flexibility as this formulation allows sampling with arbitrary resolution at inference time. PointFlow \cite{yang2019pointflow} employs continuous NFs \cite{chen2018neural} to learn two distributions in the framework of VAEs. One is the distribution of 3D shapes and another is the distribution of points given a shape. Similarly, DPF-Nets \cite{klokov2020discrete} utilizes discrete affine coupling layers \cite{dinh2016density}, resulting in a speed-up. These methods could yield point clouds that are uniformly distributed but not fine enough. Recently, MixNFs \cite{postels2021go} applies a mixtures of NFs to the decoder stage, where each flow model is used to specialize in a particular sub-cloud/part of an object. However, multiple flow models increase the computational budget in both training and inference phases.

\vspace{-3mm}
\subsection{Single Image 3D Reconstruction}

Single image 3D reconstruction aims to infer the 3D shapes of objects given a single 2D image, which is an ill-posed problem and differs from multi-view 3D reconstruction. With the development of deep learning techniques in recent years, single-view reconstruction has gained a lot of attention as one ideally expects that 3D shapes can be generated by learning from abundant single-view images. Until now, a wide range of deep generative models have been investigated from different perspectives, such as image to point cloud \cite{fan2017point, groueix2018papier, wang2019deep}, image to mesh \cite{wang2018pixel2mesh}, image to voxel \cite{klokov2019probabilistic}. To recover the lost dimension from a single image, these works train deep neural networks to learn the 2D-to-3D mapping between input images and output shapes. They typically employ an encoder-decoder structure, where the encoder maps the input image to the latent space and the decoder is supposed to perform reasoning about the 3D shape of the output space. To constrain the reconstructed 3D shapes, researchers present different supervision strategies including 3D supervision, 2D supervision, and latent supervision. Despite remarkable results achieved, the existing approaches still face many problems.

\noindent
\textbf{3D supervision:} A straightforward way is to directly compare the similarity between reconstructed 3D shape and the corresponding ground-truth 3D shape. Some works \cite{choy20163d, fan2017point, groueix2018papier, wang2019deep} employ voxel-wise cross-entropy loss or point-wise distance loss (e.g., CD, EMD). However, they may suffer from limited resolution of 3D shapes. CD and EMD are intractable for point clouds with a large number of points.

\noindent
\textbf{2D supervision:} Several works \cite{lin2018learning, navaneet2019capnet, han2020drwr, chen2021unsupervised} exploit 2D supervision, which generally leverage differentiable renderers to render a reconstructed 3D shape into 2D projections, and then minimize the point-based or pixel-based error between the projections and the corresponding 2D ground-truths. However, for the point-based loss, it is difficult to force 2D projections to uniformly cover the fine structure of ground-truths; for the pixel-based loss, it is hard to compare the images rendered from sparseness 2D projections with ground-truths due to the compactness of point clouds.

\noindent
\textbf{Latent supervision:} A few prior works combine with the latent supervision for joint optimization, which has been applied to reconstruct 3D point clouds \cite{mandikal20183d} and meshes \cite{smith2019geometrics}. They first train a 3D shape auto-encoder with a shape latent embedding, and then learn a mapping from the input image to image latent embedding which is optimized to match the corresponding shape latent embedding.

\section{Method}
\label{sec:method}
\begin{figure}[http]
\vspace{-2mm}
    \centering
    \includegraphics[width=0.79\linewidth]{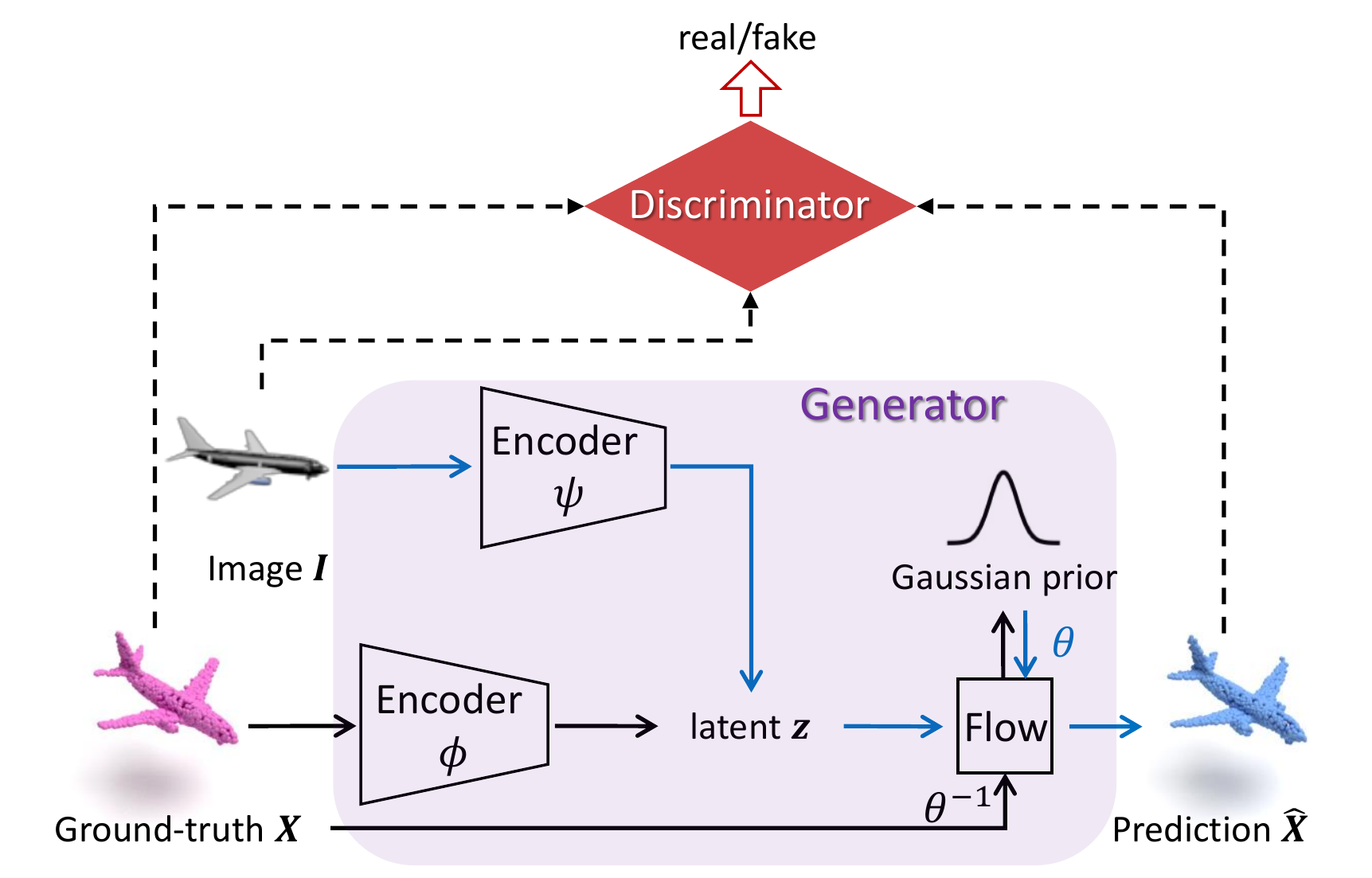}
    \caption{Overall framework. Black arrows refer to the training phase, where solid lines and dashed lines are used for generator and discriminator, respectively. Blue arrows, which are used for both training and inference, indicate the process of single-view reconstruction.}
    \vspace{-2mm}
    \label{fig:Overview}
\end{figure}

In this section, we propose a hybrid explicit-implicit generative modeling scheme for reconstructing 3D point clouds from a single 2D image. As illustrated in Fig. \ref{fig:Overview}, a new generative adversarial framework is introduced, which consists of a generator built on normalizing flows and a discriminator from cross-modal perspective. 

The basic idea is that our method takes advantages of both explicit and implicit generative modeling of point cloud generation. Using VAEs with a flow-based decoder, the generator aims to model a distribution $p(X)$ of 3D point cloud $X=\{x_1, x_2, ..., x_N\}$ where $p(x)$ denotes a point cloud over points $x\subseteq X$. Through performing maximum likelihood estimation, the generator can be regarded as an explicit density estimator which has the ability of sampling point clouds of variable size from $p(X)$ in the inference phase. On the other hand, by exploiting adversarial training strategy, the discriminator (termed as $D$) acts as an implicit sampler which targets at improving the quality of point clouds $\hat{X}$ predicted by the generator.

Prior works (e.g., \cite{klokov2020discrete, postels2021go}) leverage multiple flow models to explicitly learn the distribution of point clouds. In contrast, our hybrid explicit-implicit method has much richer representations by adding a discriminator to implicitly ensure that the generated 3D point clouds are both consistent with the input object image and exhibit a structure similar to the ground-truth point clouds. Moreover, our single-flow-based generator simplifies the two-types-of-flows pipeline used by prior works, reducing the inference runtime. For the inference, only the well-trained generator (i.e., image encoder $\psi$ and flow-based models $\theta$ shown in Fig.~\ref{fig:Overview}) is required to generate a point cloud from a single input image. 

\subsection{Flow-based Generator}
\label{sec:generator}
The generator comprises two domain-specific encoders and a decoder built on NFs. During training, a $d$-dimensional latent vector $z$ is derived using the mean $\mu_X$ and the variance $\sigma_X$ computed from $X$ by a point cloud encoder $\phi$. Hence, a distribution $q_\phi(z|X)$ is used to approximate the posterior distribution $p(z|X)$ where $z$ can be considered as an abstract shape representation of $X$. Conditioned on the shape latent $z$, a flow model containing $F$ affine coupling layers is applied to learn the invertible transformation between a simple prior distribution (e.g., Gaussian $\mathcal{N}(0,1)$) and the distribution of points (i.e., $p(X)$). This is performed by two modes: the reverse mode $\theta^{-1}$ for the transformation from $p(X)$ to $p\sim\mathcal{N}(0,1)$, and the forward mode $\theta$ for the transformation from $p\sim\mathcal{N}(0,1)$ to $p(X)$. Besides, an image encoder parameterized by $\psi$ maps the image $I$ to the latent space, which provides an image-conditioned distribution $p_\psi(z|I)$ for sampling the latent $z$ during inference. The generator can be formulated as a conditional distribution,
\begin{equation}
p(X|I)=\int_{z}p_\psi(z|I)\prod_{x\in X}p_\theta(x|z)dz
\end{equation}
The model is optimized by minimizing the negative evidence lower bound (ELBO) denoted as $\mathcal{F}$ in the following equation, 
\begin{equation}
\ln{p(X)}\geq\sum_{x\in X} \mathbb{E}_{q_\phi(z|X)}[\ln{p_\theta(x|z)}]-D_{KL}(q_\phi(z|X) \parallel p_\psi(z|I)) \equiv -\mathcal{F}
\end{equation}
where the first term is to reconstruct 3D points $x \in X$, and the second term is latent supervision which aims to reduce the KL divergence between two posterior distributions of the latent $z$. By modeling the distribution of 3D points, the flow-based generator enables flexible point cloud reconstruction from a single image, while allowing an arbitrary number of points to be sampled in the inference phase. Our generator is based on the similar VAE architecture as prior works \cite{yang2019pointflow, klokov2020discrete, postels2021go}, but differs in the use of NFs. Our method only employs a flow model for the decoding part, while previous methods adopt an additional flow model (e.g., shape flow in \cite{klokov2020discrete, postels2021go}) for modeling the latent $z$, which causes higher computational costs.

\subsection{Cross-modal Discriminator}
\label{sec:discriminator}

\begin{figure}[http]
    \centering
    \includegraphics[width=0.82\linewidth]{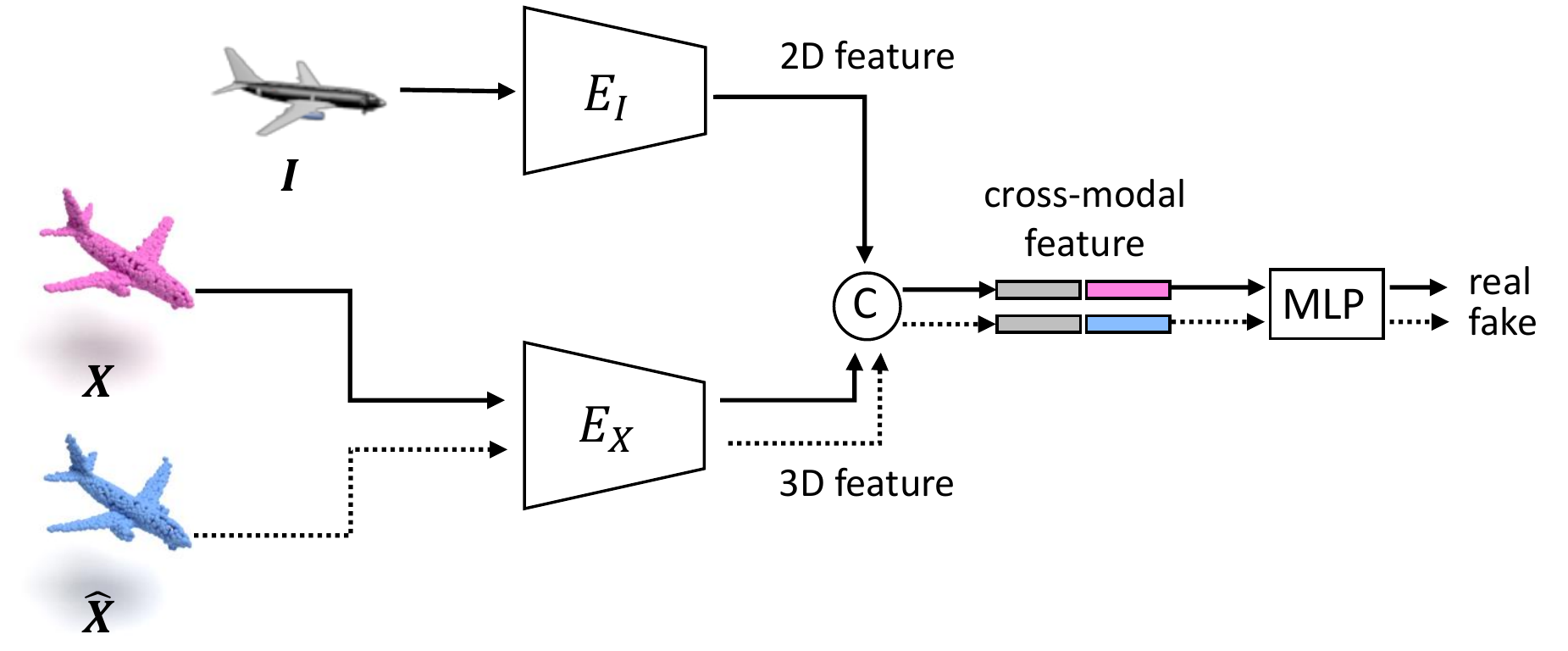}
    \caption{Architecture of the cross-modal Discriminator $D$. }
    \vspace{-3mm}
    \label{fig:Discriminator}
\end{figure}

To ensure that the generator predicts high-quality 3D point clouds, we introduce a simple but effective discriminator to regularize the prediction using adversarial training strategy. As shown in Fig. \ref{fig:Discriminator}, an encoder, $E_X$, is employed to separately extract 3D features of the predicted and ground-truth point clouds. Meanwhile, an encoder, $E_I$, is utilized to encode 2D features of the input image, which are then concatenated with the encoded 3D features. The fused cross-modality features are fed into MLP layers to output a value. Taking $I$ and $X$ as input, the value is expected as 1 (real samples); while the value is expected as 0 (fake samples) when taking $I$ and $\hat{X}$ as input. The cross-modal discriminator $D$ is trained by minimizing the least-squared loss \cite{mao2017least}, which can be formulated as follows,
\begin{equation}
\mathcal{L}_D(I,X,\hat{X})=\frac{1}{2}[(D(I,X)-1)^2+(D(I,\hat{X}))^2]+\frac{1}{2}[(D(I,\hat{X})-1)^2]
\end{equation}
\vspace{-3mm}

\section{Experiments}
In this section, we first describe the experiments, including the datasets, evaluation metrics, and experimental setup. Then, quantitative and qualitative results are given and compared with previous works. 

\noindent
\textbf{Datasets.} The experiments are conducted on ShapeNetCore.v1, a subset of the ShapeNet dataset~\cite{chang2015shapenet}, covering a variety of 3D shapes from 55 object categories. Considering that our work is concerned with single-view reconstruction, we adopt the 3D shapes of ShapeNetCore.v1 along with the corresponding 2D images rendered by 3D-R2N2 \cite{choy20163d}. For each 3D shape represented by the mesh, 3D-R2N2 \cite{choy20163d} provides $137\times137$ pixels synthesized images from 24 different viewpoints. Similar to prior works \cite{klokov2020discrete, postels2021go}, to disentangle the spatial coordinates and 3D shapes, normalized meshes are utilized to sample point clouds. For a fair comparison, we follow the train/test split rules from \cite{choy20163d}. Moreover, the proposed method is tested on  real-world images from the PASCAL3D+ dataset \cite{xiang2014beyond}.

\noindent
\textbf{Evaluation Metrics.} To quantitatively assess the reconstructed 3D point clouds, Chamfer distance (CD) \cite{fan2017point}, Earth Mover’s distance (EMD) \cite{rubner2000earth}, and F1-Score (F1) \cite{knapitsch2017tanks} are used to measure the similarity between the prediction and the corresponding ground-truth. CD computes the squared distance between each point in one set and its nearest neighbor in another set, while EMD measures the distance between two sets by attempting to transform one set into the other. For CD and EMD, a lower score signifies a better model. F1 indicates the percentage of points that are correctly reconstructed, i.e., Euclidean distance between each prediction and ground-truth under a certain threshold $\tau$. Following prior works \cite{klokov2020discrete, postels2021go}, we normalize both predicted and ground-truth point clouds into a bounding box of $[−1, 1]^{3}$, so that the metrics focus on the shape of 3D point clouds rather than the scale or position.

\noindent
\textbf{Experimental Setup.} The experiments are conducted on a single NVIDIA A40 GPU with 45-GB memory using PyTorch. We train our model for 30 epochs based on Adam optimizer \cite{kingma2014adam} and the batch size is set as 64. The initial learning rate is $2.56\times 10^{-4}$, which is then divided by 4 after  20 epochs. In the training phase, each ground-truth shape is represented by 3D point clouds with $N=2500$ points. Regarding the generator, similar to \cite{klokov2020discrete}, the point cloud encoder $\phi$ and the image encoder $\psi$ are implemented using PointNet \cite{qi2017pointnet} and ResNet18 \cite{he2016deep}, respectively. Besides, the dimension of the latent $z$ is set as $d=512$, and the NF-based decoder consists of $F=63$ affine coupling layers \cite{dinh2016density}. For the discriminator, the dimension of the 2D and 3D feature map is 128, and the cross-modality feature map obtained by concatenation operation is a 256-dimensional vector. The value is derived from 5 MLP layers. In our experiments, we implement $E_I$ using ResNet18 \cite{he2016deep}, and employ PointNet \cite{qi2017pointnet} as $E_X$. To quantitatively evaluate the performance of models, CD, EMD, and F1 with $\tau=0.001$ are reported. Mitsuba renderer \cite{Jakob2010Mitsuba} is employed for the visualization of the reconstructed 3D point clouds.

\subsection{Single-view Reconstruction Results}
During the inference phase, the latent $z$ derived from the encoding of the input test image, is used as a condition to guide the flow-based decoder in transforming the Gaussian prior distribution into a distribution of 3D points. Through sampling from such distribution, 3D point clouds with arbitrary resolutions can be reconstructed. Here we sample 2500 points for each generated point cloud to ensure comparability with prior work. 

\begin{table}[ht]
\begin{center}
\resizebox{\textwidth}{!}{%
\begin{tabular}{|c|c c c|c c c|c c c|}
\hline
\multirow{2}{*}{Methods} & \multicolumn{3}{c|}{\textit{Airplane}} & \multicolumn{3}{c|}{\textit{Car}} & \multicolumn{3}{c|}{\textit{Chair}} \\
& CD$\downarrow$ & EMD$\downarrow$ & F1$\uparrow$ & CD$\downarrow$ & EMD$\downarrow$ & F1$\uparrow$ & CD$\downarrow$ & EMD$\downarrow$ & F1$\uparrow$ \\
\hline\hline
DPF-Nets \cite{klokov2020discrete}   & 4.11 & 10.89 & 72.61 & 3.79 & 10.46 & 46.58 & 5.42 & 11.40 & 46.82\\
MixNFs \cite{postels2021go}          & 2.82 & 9.31 & 77.63 & 3.73 & 10.38 & 47.10 & 5.41 & 11.33 & 46.98 \\
Ours                                & \textbf{2.33} & \textbf{8.68} & \textbf{79.94} & \textbf{3.60} & \textbf{10.24} & \textbf{47.71} & \textbf{5.02} & \textbf{10.99} & \textbf{49.09}\\ \hline
\textit{Oracle}                              & 0.50 & 4.48 & 97.62 & 1.55 & 6.34 & 73.65 & 1.11 & 5.92 & 84.32\\
\hline
\end{tabular}%
}
\end{center}
\caption{Results on three categories of the ShapeNetCore.v1 dataset~\cite{chang2015shapenet}. CD is multiplied by $10^3$; EMD is multiplied by $10^2$; F1 is the \% value. The best results are shown in \textbf{bold}.}
 \vspace{-1mm}
\label{table:single_class}
\end{table}

\begin{figure}[t]
    \centering
    \includegraphics[width=\linewidth]{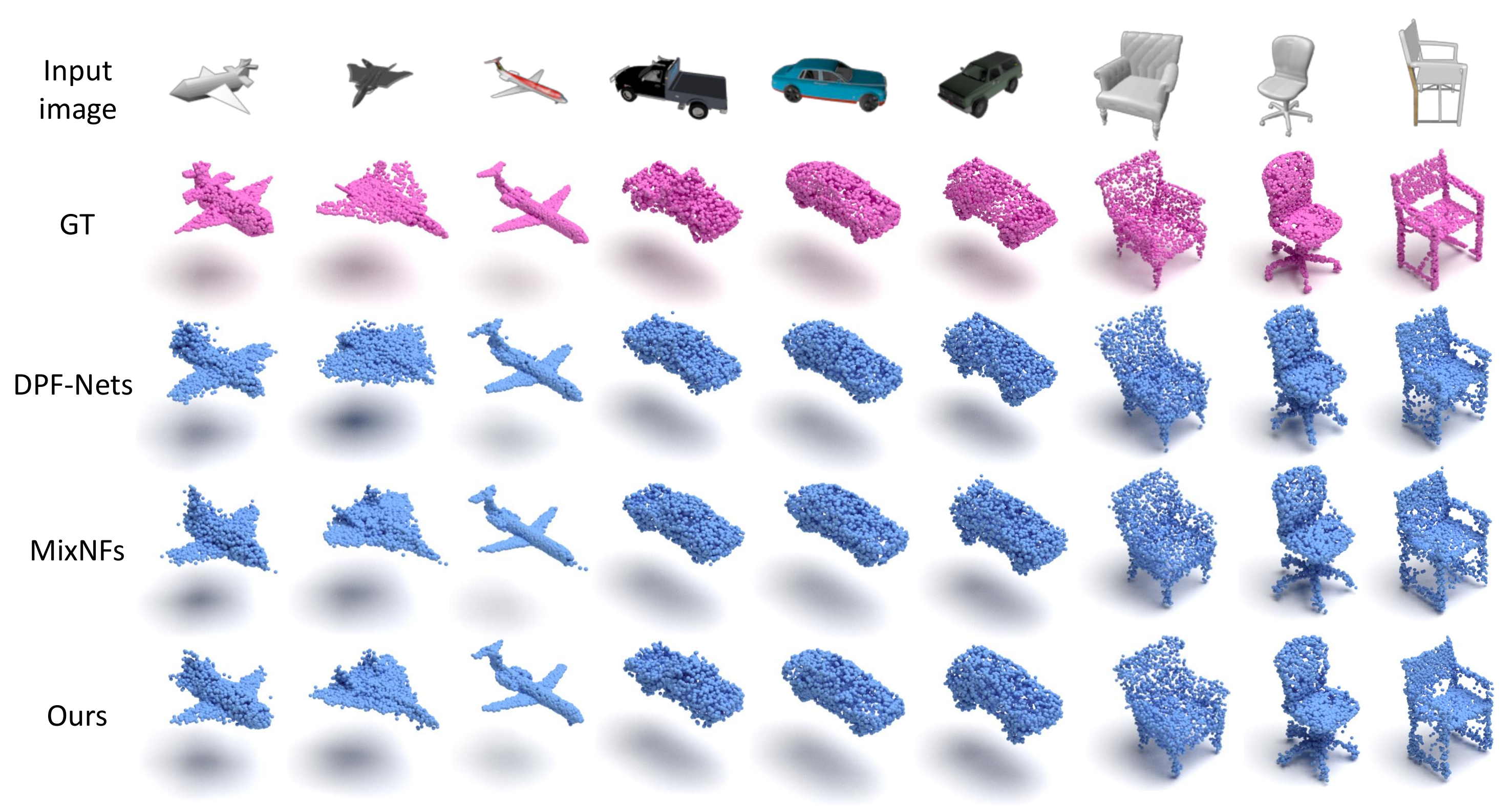}
    \caption{Qualitative results on the \textit{Airplane}, \textit{Car}, and \textit{Chair} of the ShapeNetCore.v1 dataset.}
    \vspace{-3mm}
    \label{fig:results}
\end{figure}

First, our method is compared with recent similar works \cite{klokov2020discrete, postels2021go} on three major categories (i.e., \textit{Airplane}, \textit{Car}, \textit{Chair}) of the ShapeNetCore.v1 dataset. Table \ref{table:single_class} lists the results w.r.t. CD, EMD and F1, where all the models are trained and evaluated given a single object category. By directly sampling another point cloud from the ground-truth mesh without training, we also provide an "\textit{Oracle}" to show the best possible performance on the experimental dataset. It can be observed that  our method outperforms the baseline methods DPF-Nets \cite{klokov2020discrete} and MixNFs \cite{postels2021go} on these three categories. In addition, some results are illustrated in Fig. \ref{fig:results}. Compared to DPF-Nets \cite{klokov2020discrete}, MixNFs \cite{postels2021go} improves the local details by adopting multiple flow models to decode different object parts, but sometimes, it still fails to reconstruct fine 3D structures for complicated shapes. In contrast, the 3D point clouds produced by our method demonstrate cleaner global shapes (e.g., the airplanes in the first three columns) conditioned on the input images as well as finer local structures (e.g., the car’s rear-view mirror in the sixth column and the chair’s castor wheel in the eighth column) close to the ground-truths.

\begin{figure}[ht]
    \centering
    \includegraphics[width=0.76\linewidth]{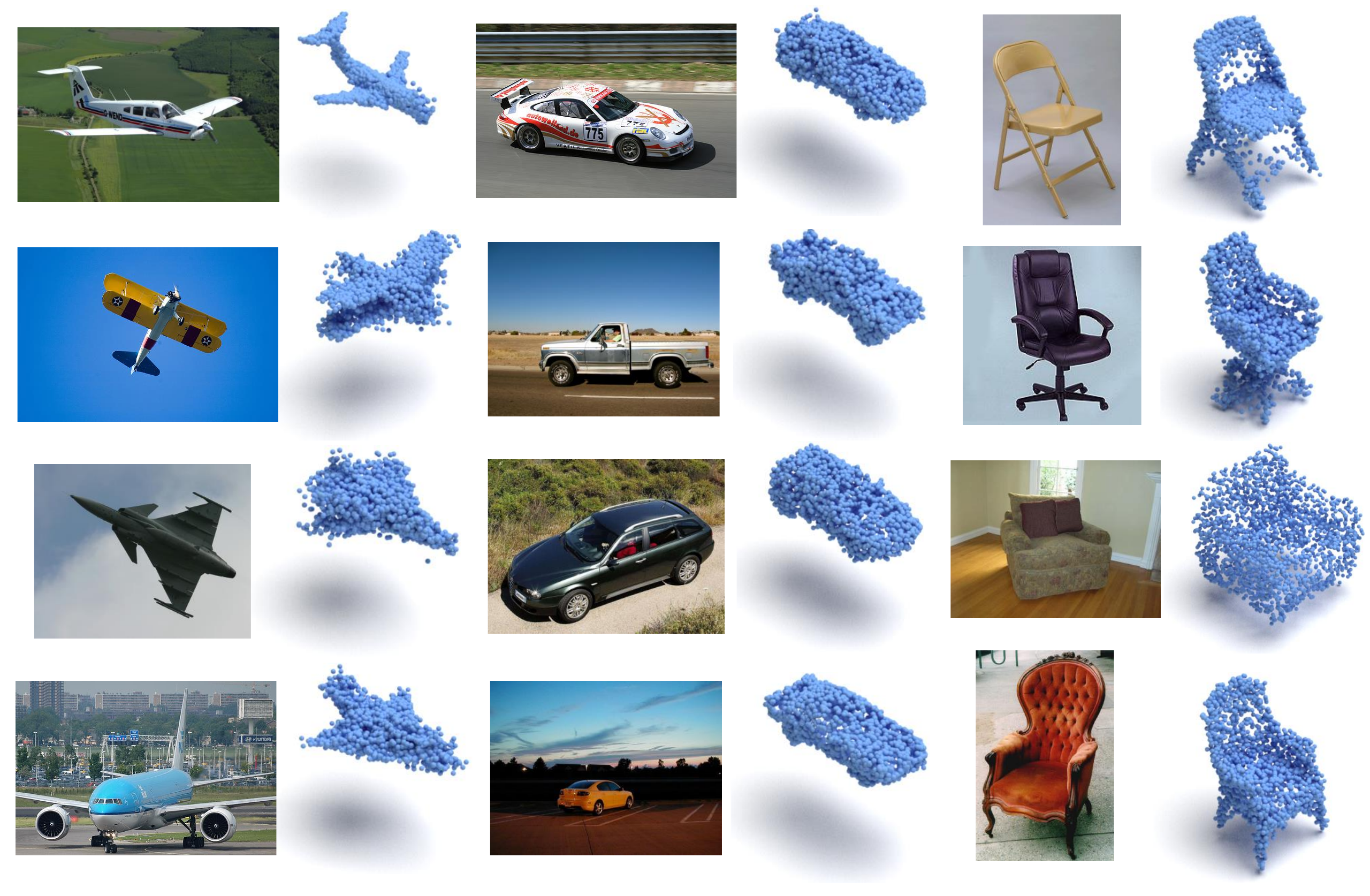}
    \caption{Qualitative results on the \textit{Airplane}, \textit{Car}, and \textit{Chair} of the PASCAL3D+ dataset.}
    \vspace{-3mm}
    \label{fig:realimg}
\end{figure}

Fig.~\ref{fig:realimg} shows the results of our method on real images from the PASCAL3D+ dataset \cite{xiang2014beyond}, which is only used for testing the models pre-trained on certain single category of the ShapeNetCore.v1 dataset. Given a single real image containing a single object-of-interest, our method can obtain its realistic 3D point cloud. This demonstrates the generalization capacity of our method since the 3D shapes can be reconstructed from real images even if the models are trained on synthetic images. Besides, as can be seen from the last row, there might be a mismatch between the generated shape and the input real image due to the complexity of the real image itself and the gap between real and synthetic images. 

\begin{table}[ht]
\begin{center}
\begin{tabular}{|c|c c c c|}
\hline
Methods & CD$\downarrow$ & EMD$\downarrow$ & F1$\uparrow$ & Speed$\uparrow$ \\
\hline\hline
PRN \cite{klokov2019probabilistic}   & 7.56 & \textbf{11.00} & \textbf{53.1} & -\\
AtlasNet \cite{groueix2018papier}    & {5.34} & 12.54 & 52.2 & -\\
DCG \cite{wang2019deep}              & 6.35 & 18.94 & 45.7 & -\\
Pixel2Mesh \cite{wang2018pixel2mesh} & 5.91 & 13.80 & - & -\\
DPF-Nets \cite{klokov2020discrete}   & 5.55 & {11.11} & 51.7 & {259}\\
MixNFs \cite{postels2021go}          & 5.66 & 11.20 & 52.3 & 5\\
Ours                                & \textbf{5.32} & \textbf{11.00} & {53.0} & \textbf{273}\\ \hline
\textit{Oracle}                               & 1.10 & 5.70 & 84.0 & -\\
\hline
\end{tabular}
\end{center}
\caption{Comparative results on 13 categories of the ShapeNetCore.v1 dataset~\cite{chang2015shapenet}. Speed is the number of samples processed per second.  The best results are shown in \textbf{bold}. }
 \vspace{-1mm}
\label{table:all_class}
\end{table}

To further verify the generalization ability of the proposed method, we train and evaluate on all 13 categories of the intersection of ShapeNetCore.v1 and 3D-R2N2. Our method is compared with related approaches on single-view reconstruction in Table \ref{table:all_class}. These include early methods PRN \cite{klokov2019probabilistic}, AtlasNet \cite{groueix2018papier}, DCG \cite{wang2019deep} and Pixel2Mesh \cite{wang2018pixel2mesh}, as well as recent methods DPF-Nets \cite{klokov2020discrete} and MixNFs \cite{postels2021go}. Specifically, results of recent methods are obtained by using the official implementation provided by the authors, while results of early methods are obtained from the original paper \cite{klokov2020discrete}. It can be seen that our method achieves the best results w.r.t. CD and EMD, while very close to the best result w.r.t. F1 score. Besides, the inference speed is provided for our method and the recent methods. 

\subsection{Ablation Studies}

For the ablation studies, we consider how the following aspects influence the final performance of the proposed method.

\begin{table}[http]
\setlength\tabcolsep{.37em}
\renewcommand{\arraystretch}{1.1} 
\begin{center}
\begin{tabular}{|c|c|c c c|}
\hline
Generator & Discriminator & CD$\downarrow$ & EMD$\downarrow$ & F1$\uparrow$ \\
\hline\hline
w Shape Flow & w/o & 5.55 & 11.11 & 51.7\\
w/o Shape Flow & w/o & 5.50 & 11.20 & 52.3\\
w/o Shape Flow & w/ $D$ & \textbf{5.32} & \textbf{11.00} & \textbf{53.0} \\
\hline
\end{tabular}
\end{center}
\caption{Ablative results on 13 categories of the ShapeNetCore.v1 dataset~\cite{chang2015shapenet} w.r.t. different architectures. The best results are shown in \textbf{bold}. }
 \vspace{-1mm}
\label{table:architecture}
\end{table}

\noindent
\textbf{Comparison of Different Architectures.} 
To a certain extent, the baseline methods DPF-Nets \cite{klokov2020discrete} and MixNFs \cite{postels2021go} share a similar pipeline for single-view reconstruction. They employ two types of flows: a shape flow for invertible transformation between the distribution of shape latent and the distribution of image latent, and a conditional flow for invertible transformation between the Gaussian prior distribution and the distribution of 3D points. The difference between DPF-Nets and MixNFs lies in the decoding stage where DPF-Nets applies a single conditional flow while MixNFs applies multiple conditional flows to process different object parts separately. As described in Sec. \ref{sec:method}, our model consists of a generator and a discriminator. In contrast, our single-flow-based generator simplifies the two-types-of-flows pipeline used by \cite{klokov2020discrete, postels2021go}, and our discriminator is only utilized for training, which reduces the computational costs suffered by \cite{postels2021go}. It can be observed from Table \ref{table:architecture} that our model (i.e., the last row) achieves the best performance compared to other architectures in different component settings.

\begin{table}[http]
\begin{center}
\resizebox{\textwidth}{!}{%
\begin{tabular}{|c|c c c|c c c|c c c|}
\hline
\multirow{2}{*}{Methods} & \multicolumn{3}{c|}{$N=1024$} & \multicolumn{3}{c|}{$N=2500$} & \multicolumn{3}{c|}{$N=4096$} \\
& CD$\downarrow$ & EMD$\downarrow$ & F1$\uparrow$ & CD$\downarrow$ & EMD$\downarrow$ & F1$\uparrow$ & CD$\downarrow$ & EMD$\downarrow$ & F1$\uparrow$ \\
\hline\hline
DPF-Nets \cite{klokov2020discrete}   & 7.42 & 12.46 & 34.6 & 5.55 & 11.11 & 51.7 & 4.94 & 10.62 & 60.0\\
MixNFs \cite{postels2021go}          & 7.50 & 12.55 & 35.2 & 5.66 & 11.20 & 52.3 & 5.05 & 10.71 & 60.5 \\
Ours                                & \textbf{7.15} & \textbf{12.37} & \textbf{35.6} & \textbf{5.32} & \textbf{11.00} & \textbf{53.0} & \textbf{4.72} & \textbf{10.51} & \textbf{61.3}\\ \hline
\textit{Oracle}                              & 2.45 & 7.96 & 60.5 & 1.10 & 5.70 & 84.0 & 0.70 & 4.74 & 92.7\\
\hline
\end{tabular}%
}
\end{center}
\caption{Ablative results on 13 categories of the ShapeNetCore.v1 dataset~\cite{chang2015shapenet} w.r.t. different number of points. The best results are shown in \textbf{bold}.}
 \vspace{-1mm}
\label{table:numpoints}
\end{table}

\noindent
\textbf{Comparison of Sampling Different Number of Points.} Since we employ the flow-based decoder in the generator, our method inherits the flow-based generative methods for sampling an arbitrary number of points. Table \ref{table:numpoints} shows the performance of different methods when sampling 1024, 2500 and 4096 points for each 3D shape. It can be seen that the performance improves as the number of sampling points increases. The proposed method outperforms other methods, demonstrating the effectiveness and stability of our method.

\begin{table}[ht]
\begin{center}
\begin{tabular}{|c|c c c |}
\hline
Methods & CD$\downarrow$ & EMD$\downarrow$ & F1$\uparrow$ \\
\hline\hline
DPF-Nets \cite{klokov2020discrete}   & $5.553 \pm 0.0005$ & $11.111 \pm 0.0019$ & $51.66 \pm 0.0007$ \\
MixNFs \cite{postels2021go}          & $5.660 \pm 0.0010$ & $11.200 \pm 0.0015$ & $52.30 \pm 0.0000$ \\
Ours                                & $\textbf{5.321} \pm 0.0004$ & $\textbf{10.999} \pm 0.0009$ & $\textbf{52.99} \pm 0.0017$ \\ 
\hline
\end{tabular}
\end{center}
\caption{Ablative results on 13 categories of the ShapeNetCore.v1 dataset~\cite{chang2015shapenet} w.r.t. different samplings with the same number of points (2500). Note that the mean and standard deviation are shown here for five sampling runs. The best results are shown in \textbf{bold}.
}
\vspace{-1mm}
\label{table:differsample}
\end{table}

\noindent
\textbf{Comparison of Different Samplings With The Same Number of Points.} Table \ref{table:differsample} shows the results of different methods (mean and  standard deviation) when sampling five times with 2500 points for each 3D shape. It is observed that different samplings result in very similar scores w.r.t. all three metrics, suggesting that it does not affect the stability of results if another sampling with the same number of points is chosen.

\section{Conclusion}

In this paper, we propose a hybrid explicit-implicit generative modeling scheme for 3D point cloud reconstruction from a single image. To solve the limitation caused by generating point clouds with fixed resolution, we introduce a single-flow-based generator to approximate the distribution of 3D points, which allows us to sample an arbitrary number of points. Besides, a cross-modal discriminator is developed to guide the generator to produce high-quality point clouds, which are both reasonable conditioned on the input image and exhibit a 3D structure similar to the ground-truth. The effectiveness and generalization ability of our method has been demonstrated by the experimental results on ShapeNet and PASCAL3D+ datasets. 


\clearpage
\bibliography{egbib}
\end{document}